\documentclass[letterpaper]{article} 
\usepackage{aaai24}  
\usepackage{times}  
\usepackage{helvet}  
\usepackage{courier}  
\usepackage[hyphens]{url}  
\usepackage{graphicx} 
\usepackage{natbib}  
\usepackage{caption} 
\usepackage{algorithm}
\usepackage{algorithmic}
\usepackage{amsmath}
\usepackage{multirow} 
\usepackage{tabularray}
\usepackage{subcaption}
\usepackage{longtable}
\newcommand{\methodname}{{\tt{LR-XFL}}}

\usepackage{newfloat}
\usepackage{listings}
\DeclareCaptionStyle{ruled}{labelfont=normalfont,labelsep=colon,strut=off} 
\floatstyle{ruled}
\newfloat{listing}{tb}{lst}{}
\floatname{listing}{Listing}
\title{\methodname{}: Logical Reasoning-based Explainable Federated Learning}

\author {
    Yanci Zhang,
    Han Yu
}
\affiliations {
    School of Computer Science and Engineering, Nanyang Technological University, Singapore\\
    Joint NTU-WeBank Research Centre on Fintech, Nanyang Technological University, Singapore\\
    \{yanci001, han.yu\}@ntu.edu.sg
}

\usepackage{bibentry}

\begin{document}

\maketitle

\begin{abstract}

Federated learning (FL) is an emerging approach for training machine learning models collaboratively while preserving data privacy. The need for privacy protection makes it difficult for FL models to achieve global transparency and explainability. To address this limitation, we incorporate logic-based explanations into FL by proposing the \underline{L}ogical \underline{R}easoning-based e\underline{X}plainable \underline{F}ederated \underline{L}earning (\methodname{}) approach. Under \methodname{}, FL clients create local logic rules based on their local data and send them, along with model updates, to the FL server. The FL server connects the local logic rules through a proper logical connector that is derived based on properties of client data, without requiring access to the raw data. In addition, the server also aggregates the local model updates with weight values determined by the quality of the clients' local data as reflected by their uploaded logic rules. The results show that \methodname{} outperforms the most relevant baseline by 1.19\%, 5.81\% and 5.41\% in terms of classification accuracy, rule accuracy and rule fidelity, respectively. The explicit rule evaluation and expression under \methodname{} enable human experts to validate and correct the rules on the server side, hence improving the global FL model's robustness to errors. It has the potential to enhance the transparency of FL models for areas like healthcare and finance where both data privacy and explainability are important.

\end{abstract}

\section{Introduction}

Federated learning (FL)~\cite{Kairouz-et-al:2021} is a collaborative training paradigm that jointly trains artificial intelligence (AI) models from a set of FL clients without exposing their local data. Under the guidance of a central server, clients improve their local models from the information gained from other clients. The general idea is that clients upload their model updates rather than data to the FL server. The server aggregates the received models and returns the updated global model to clients so that they can further train it. This privacy-protection joint training schema, however, introduces complexities, particularly in explaining the decision-making process, given the server's lack of direct access to client datasets.

Explainable AI (XAI)~\cite{gunning2019xai,xu2019explainable,Yu-et-al:2014AAMAS} is gaining attention in recent years. The aim of XAI is to make AI model behaviours interpretable to humans by providing explanations. A wide range of studies have successfully provided explanations for black-box AI models. For instance, approaches like Grad-CAM~\cite{selvaraju2017grad} analyse the output of the final convolutional layer of a given neural network and calculate how changes to each region of an input image affect the output to identify the most important image regions for decision-making. Model-agnostic algorithms like SHAP~\cite{lundberg2017unified} and LIME~\cite{ribeiro2016should} assess the importance of features at the instance level. However, these approaches cannot clearly illustrate the model decision-making processes, but rather provide a post-hoc analysis of the models. 

Concept-based models, as a bridge between symbolic AI (based on rules and logic) and statistical AI (e.g., neural networks), extract concepts from the model by mapping the hidden information of the last layer of neurons in a neural network to human-understandable concepts~\cite{kim2018tcav} or constraining concepts as a part of a neural network~\cite{koh2020concept}. However, the concepts are still isolated and at the instance level. Logic rules, however, can connect the activated concepts during the prediction to illustrate the reasoning process~\cite{lee2022self}. Instance-level logic rules can be further connected to form a global class-level rule~\cite{barbiero2022entropy}. Such logic-based explanations are beneficial as they illustrate the decision-making process in a manner that is readily understood by humans.

Traditional logic-based concept models are designed for centralised AI frameworks and cannot be directly applied in FL settings. Integrating logic rules into FL faces three important challenges: 
\begin{enumerate}
    \item \textbf{Local Accuracy vs. Global Representativeness}: Clients, with access to only partial or potentially skewed data, may derive rules that seem accurate within their local context but could introduce biases or misrepresent the broader global perspective. It is imperative for the server to generate comprehensive and globally accurate rules. 
    \item \textbf{Conflict Resolution and Rule Combination}: The merging of logic rules at a central server can lead to conflicts. A simplistic connection of local rules using the logical `AND' operator might introduce conflicts, and combining rules with the `OR' operator could dilute the overall rule accuracy. Moreover, determining the optimal combination of rules remains a formidable challenge. Global rules should selectively incorporate accurate rules, using appropriate logical connectors.
    \item \textbf{FL Client Weight Assignment}: Assigning suitable weights to client model updates during aggregation based on their logic rules presents another challenge.
\end{enumerate}

To tackle these challenges, we propose the \underline{L}ogical \underline{R}easoning-based e\underline{X}plainable \underline{F}ederated \underline{L}earning (\methodname{}) approach. Under \methodname{}, FL clients create local logic rules based on their local data and send them, along with model updates, to the FL server. The FL server connects the local logic rules through a logical connector (AND or OR) that is adaptively determined by \methodname{} based on properties of client data, without exposing raw data. In addition, the server also aggregates the local model updates with weight values determined by the quality of the clients' local data as reflected by their uploaded logic rules. To the best of our knowledge, \methodname{} is the first FL reasoning approach capable of adaptively aggregating local rules from clients.

To evaluate the performance of \methodname{}, we conduct extensive experiments on four benchmark datasets under FL settings\footnote{The code is available at \url{https://github.com/Yanci87/LR-XFL}.}. Compared to three prevailing related alternative approaches, \methodname{} has demonstrated significant advantages. It outperforms the most relevant baseline by 1.19\%, 5.81\% and 5.41\% in terms of classification accuracy, rule accuracy and rule fidelity, respectively. 
The explicit rule evaluation and expression under \methodname{} enables human experts to validate and correct the rules on the server side, hence improving the global FL model's robustness to errors. It has the potential to enhance the transparency of FL models for areas like healthcare and finance where both data privacy and explainability are important.

\section{Related Work}
\subsection{Concept-based XAI}
Concept-based XAI focuses on identifying high-level abstractions or ``concepts'' in data. In traditional deep learning models, the lower layers often detect edges or textures, while the upper layers detect more abstract features like shapes or objects. Concept-based learning regards the abstract features as ``concepts'' and aims to make them more explicit. The concepts are often obtained by linking the hidden information of the last layer of a neural network to human-understandable concepts~\cite{kim2018tcav, kazhdan2020now}, or constraining the structure of the neural network to learn the concepts~\cite{chen2020concept, ciravegna2020human, koh2020concept, stammer2021right,barbiero2022entropy}. The design of a concept-based model makes it suitable to provide explanations for the decision-making process as the concepts are intuitive and understandable for humans, thereby making the decision process transparent. For instance, if a model trained on animal images learns the concept of ``wings'' and ``beaks'', it can explain a classification decision by reasoning that the presence of wings in an image is a strong indicator for the ``bird'' category. 

Logic rules serve as a means to link extracted concepts, enabling the decisions made by the model to be explained based on the concepts it incorporates. Logic rules are inherently straightforward and deterministic. Traditional rule-based systems, such as Decision Trees (DTs)~\cite{quinlan1986induction}, offer intuitive explanations through logic rules. In DTs, each decision path can be interpreted as a rule. Nonetheless, including all features in a decision path can render a rule excessively lengthy with unnecessary features, and consequently diminish its comprehensibility. In the domain of natural language processing, logical reasoning has been adopted for tasks like sentiment analysis~\cite{lee2022self} and text prediction~\cite{jain-etal-2022-extending}. For image and tabular data, a novel entropy-based linear layer has been introduced to produce rule-based explanations~\cite{barbiero2022entropy}. However, these approaches require direct access to the training data, making them unsuitable for operations under FL settings.

\subsection{Federated Learning}
FL offers a decentralised and privacy-preserving approach to training AI models on distributedly owned data~\cite{Kairouz-et-al:2021}. Instead of moving potentially sensitive raw data, only model updates are being sent back and forth between the FL server and the FL clients. However, such a training approach makes it challenging for explaining the decision-making rationale behind the resulting FL model. The FL field has recognised the potential for logic rules to address this challenge. There have been several notable attempts to integrate logic rules into FL. 

In~\cite{an2023guiding}, signal temporal logic is employed to discern the properties of client devices, which are subsequently leveraged to cluster them during model aggregation to produce personalised FL models. Similarly in~\cite{cha2022fuzzy}, fuzzy logic is used to enhance client selection in vehicular networks. Besides, in~\cite{zhu2021horizontal}, the Takagi–Sugeno fuzzy rule is integrated into FL for federated fuzzy clustering. Nevertheless, these approaches are predominantly focused on FL client selection and clustering. They are not designed to address the challenges of building explainable FL with logical reasoning. The proposed \methodname{} approach bridges this important gap.

\section{Preliminaries}
The base model under the proposed \methodname{} approach is the entropy-based logic explanations of neural networks~\cite{barbiero2022entropy}. This model adeptly extracts logic-based explanations from neural networks, representing them in logic rules. Designed to handle both images and tabular data, the model provides classification together with rule-based explanations.
For image data, it first employs the ResNet10~\cite{he2016deep} image processing network to map a given image from the pixel space to the concept space. For tabular data, the mapping can be achieved by linear models. Subsequently, the concept space embedding is mapped to the target class using the entropy-based linear layer. The activated concepts are then derived from the parameters of the entropy-based layer to form logic rules.

The process of obtaining logic rules from the entropy-based linear layer relies on a truth table, denoted as $T_c$, corresponding to class $c$. This truth table $T_c$ captures the behaviour of the neural network by leveraging Boolean-like representations of the input concepts. Specifically, each row of $T_c$ encompasses activated concepts for a sample predicted to be under class $c$. The activation status of these concepts for a single data point is determined by processing its concept vector via a binary mask derived from the entropy-based layer parameters. Concepts that are activated through the binary mask for the prediction of class $c$ are included, while others are excluded. Given an activated concept $f_i$, its representation will either be $f_i$ or $\neg f_i$, depending on its value within the input data point. For every row in the truth table, a sample-level rule-based explanation is formulated by connecting all activated concepts using the AND operator. The class-level explanation for class $c$ is obtained by connecting these sample-level rules, pertaining to data classified under class $c$, with the OR operator.

However, always using the OR operator to connect sample-level rules can inadvertently make the rule less precise. Consider two sample-level rules: 1) $hasBeak \leftrightarrow Bird$, and 2) $hasWing \leftrightarrow Bird$. Combining these rules with the OR operator yields: $hasBeak \vee hasWing \leftrightarrow Bird$. However, this aggregated rule incorrectly classifies samples only with beaks and no wings, or only with wings and no beaks as birds. In this case, the AND operator is more suitable since the two sample-level rules contain complementary information from different perspectives. The proposed \methodname{} approach is designed to deal with such challenging situations.

\begin{figure}[!b]
    \centering
    \includegraphics[width=0.77\columnwidth]{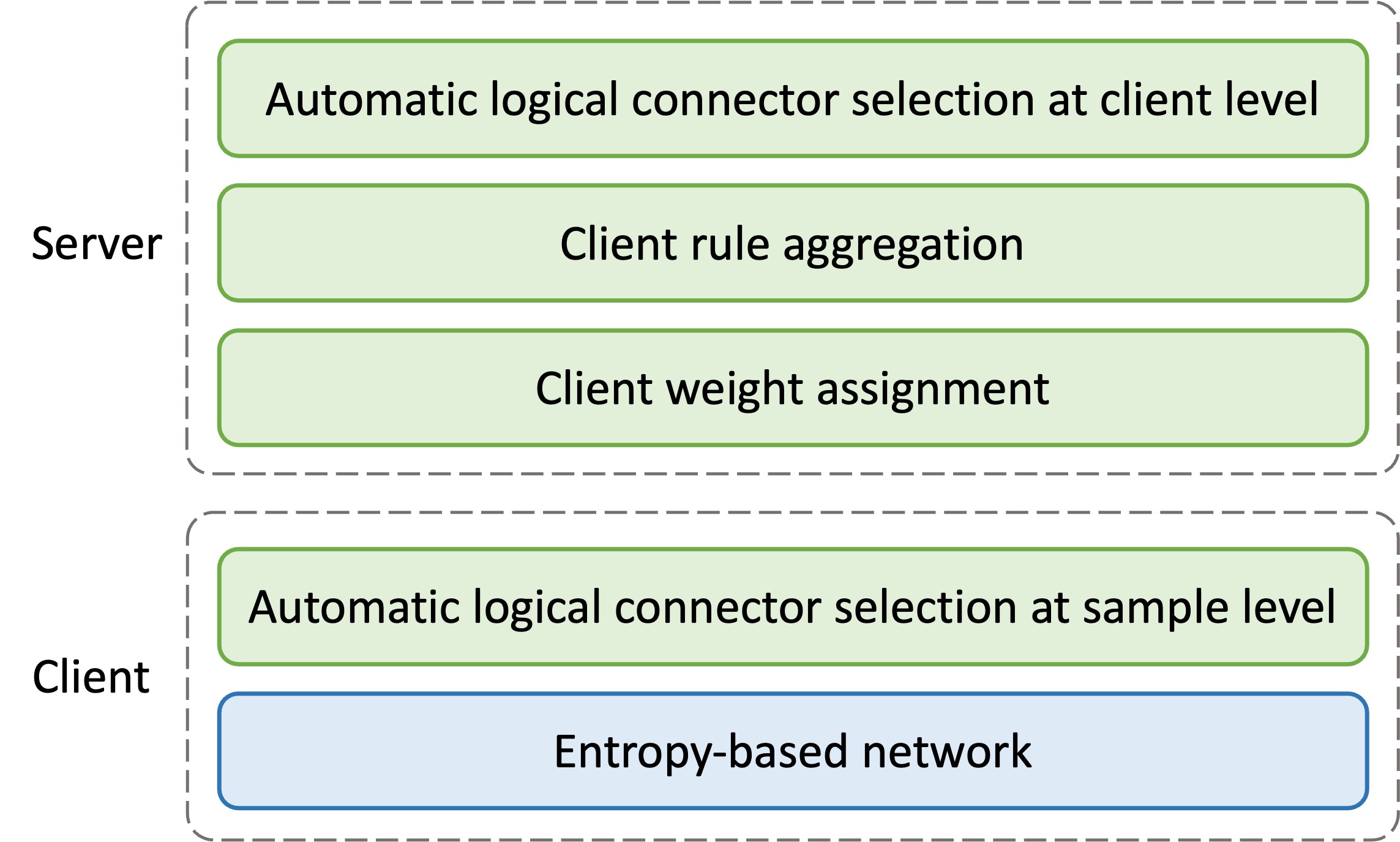}
    \caption{Overall design of \methodname{}.}
    \label{fig:block}
\end{figure}

\section{The Proposed \methodname{} Approach}
In this section, we describe the detailed design of \methodname{} (Figure~\ref{fig:block}), a first-of-its-kind logic-based explainable framework designed for FL settings built upon the entropy-based network~\cite{barbiero2022entropy}. It consists of a novel method for automatically determining the appropriate logical connector (AND or OR). In addition, it addresses the challenge of rule aggregation when selecting and merging client-generated rules. During FL model aggregation, \methodname{} computes client weights based on their respective rules to support weighted averaging.

\subsection{Overview of \methodname{}}
\begin{figure*}[!t]
    \centering
    \includegraphics[width=0.85\textwidth]{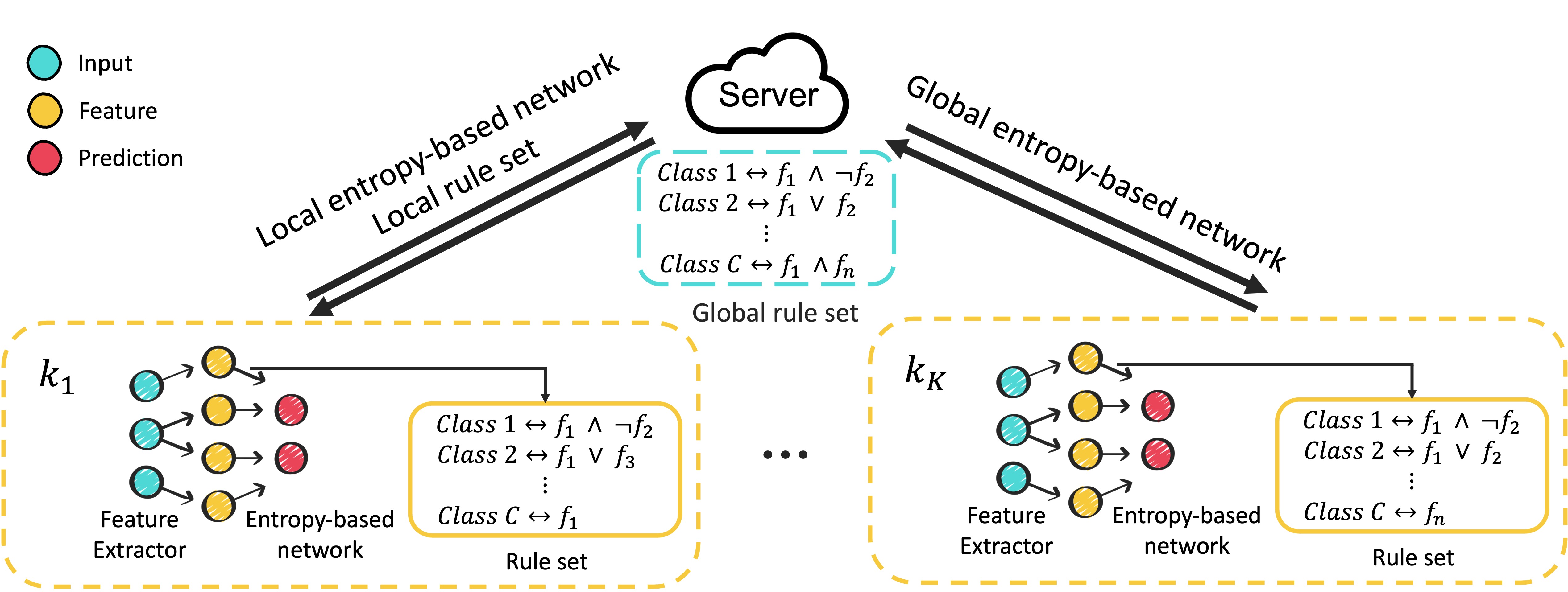}
    \caption{The system architecture and workflow of \methodname{}.}
    \label{fig:workflow}
\end{figure*}

Figure~\ref{fig:workflow} depicts the architecture and workflow of \methodname{}. On the client side, entropy-based networks in~\cite{barbiero2022entropy} are adopted as the base model. Each FL client possesses a set of rules derived from this entropy-based network. The FL server derives a set of global rules from local rules, and is in charge of sending the global model back to clients.
Algorithm~\ref{alg:algorithm} provides an overview of \methodname{}. Under \methodname{}, FL clients formulate local logic rules grounded in their individual datasets. These rules, along with model updates, are then transmitted to the FL server. Notably, the FL server aggregates the local logic rules through an appropriate connector ($\wedge$ or $\vee$), which is determined based on the characteristics inherent in the client's data without accessing the raw data. Furthermore, the server aggregates local model updates, assigning weight values based on the quality of the clients' local data as gauged from their local logic rules. This iterative training process is carried out until either a predefined maximum number of iterations $T$ is reached, or the global model attains its target performance on the server-side validation dataset.

\begin{algorithm}[!t]
\caption{\methodname{}}
\label{alg:algorithm}
\textbf{Input}: $K$ clients, each holding a set of local data; a server, holding a set of data for logic validation and testing\\
\textbf{Output}: Global logic rules for the server; local models and logic rules for clients
\begin{algorithmic}[1] 
\WHILE{Global model has not achieved the target performance on the validation set \AND max training rounds have not been reached}
    \STATE \textbf{For each FL client $k, k\in\{1, \dots ,K\}$}:
        \STATE Trains the local model;            
        \STATE Selects the appropriate logical connector for the local rules;
        \STATE Generates logic rules $r_{k}^{c}$ for $c\in\{1,\dots, C\}$ classes;
        \STATE Uploads the local model and logic rules to the FL server;
    \STATE \textbf{FL Server}:
        \STATE Selects the appropriate global logical connector based on clients' uploaded logic rules;
        \STATE Selects and aggregates clients' local rules;
        \STATE Calculates and assigns weights $\{w_{1}, \dots, w_{K}\}$ for the clients based on the performance of their logic rules;
        \STATE Aggregates the local models into the global model based on the assigned weights;
        \STATE Sends the global model to the clients;
    \STATE \textbf{For each FL client $k$}: Receives the global model and continues training for the next round;
\ENDWHILE
\end{algorithmic}
\end{algorithm}

\subsection{Determining OR or AND Logical Connector}
Ideally, the global rules shall only include the right rules with the correct logical connector. When integrating local logic rules, the first step is to determine whether to use the AND or the OR logical connector. Connecting local rules with AND might lead to conflicts, while introducing partial or wrong rules with OR might undermine accuracy.

We posit that the underlying rationale for choosing between AND and OR hinges on the potential conflict among the clients' local features. If all features are mutually exclusive (i.e., they cannot coexist in a single data point), the $\vee$ connector shall be used. An example of this scenario is the classification sequence $1 \vee 3 \vee 5 \vee 7 \vee 9$, which denotes the `odd number' category. On the other hand, in instances where features do not inherently conflict and can appear together in a single data point, the $\wedge$ connector is preferable. For instance, in an animal classification task, features ``wings" and ``beaks" can co-exist in the ``bird" category. Therefore, a rule such as $Wings \wedge Beaks \leftrightarrow Birds$ is suitable for classifying birds.

To determine the appropriate connector, we designed a positive co-occurrence matrix and a negative co-occurrence matrix to explore possible feature conflicts. The co-occurrence matrix captures the occurrence of pairs of features together in a certain rule. Specifically, consider the sample-level rule set $ R_{k}^{c} $ for class $ c $ in client $ k $, where each rule $ r $ is defined as $ f_{1} \wedge \neg f_{2} \wedge \dots \wedge f_{n} $:

\begin{enumerate}
    \item \textbf{The Positive Co-occurrence Matrix} 
    records the number of times features $f_{i}$ and $f_{j}$ appearing together in rules in the form of $f_{i} \wedge \dots \wedge f_{j} $. If features $f_{i}$ and $f_{j}$ jointly appear in rules for $t$ times, $p_{ij} = p_{ji} = t$.
    \begin{equation}
        M_{pos} = 
    \begin{bmatrix}
    p_{11} & \cdots & p_{1n} \\
    \vdots & \ddots & \vdots \\
    p_{n1} & \cdots & p_{nn}
    \end{bmatrix}. \label{eq:1}
    \end{equation}

    \item \textbf{The Negative Co-occurrence Matrix} records the number of times features $f_{i}$ and $f_{j}$ appear together in rules in the form of $f_{i} \wedge \dots \wedge \neg f_{j} $. If features $f_{i}$ and $\neg f_{j}$ appear in rules for $t$ times, $q_{ij} = t$. But unlike the positive co-occurrence matrix, in the negative co-occurrence matrix, $q_{ij} \neq q_{ji}$. $q_{ij}$ records the number of times $f_{i}$ and $\neg f_{j}$ appear in a rule, while $q_{ji}$ records the co-occurrence of $f_{j}$ and $\neg f_{i}$.
    \begin{equation}
        M_{neg} = 
    \begin{bmatrix}
    q_{11} & \cdots & q_{1n} \\
    \vdots & \ddots & \vdots \\
    q_{n1} & \cdots & q_{nn}
    \end{bmatrix}. \label{eq:2}
    \end{equation}
    
\end{enumerate}

We proposed two criteria to evaluate the extent of feature conflict through the positive co-occurrence matrix $M_{pos}$ and the negative co-occurrence matrix $M_{neg}$.

\begin{enumerate}
    \item \textbf{Diagonality}, which is calculated as: 
    \begin{equation}
        D = \frac{\sum_{i=1}^{n} p_{ii}}{\sum_{i=1}^{n} \sum_{j=1}^{n} p_{ij}} \label{eq:3}
    \end{equation}
    where $n$ is the total number of features.
    A high diagonality suggests that a feature is likely to appear in a rule with no other features than itself (i.e., a high likelihood of feature conflict).
    
    \item \textbf{Exclusivity}, which is calculated as:
    \begin{equation}
        E = \frac{\max_{i=1}^{n} \left( \sum_{j=1}^{n} q_{ij} \right)}{\sum_{m=1}^{M} l_{r_{m}}} \label{eq:4}
    \end{equation}
    where $l_{r_{m}}$ is the length of rule $r_{m}$ and $M$ is the total number of rules contributing to the negative co-occurrence matrix $M_{neg}$. The exclusivity $E$ measures the average feature negative co-occurrence in a client. A high exclusivity indicates that a rule is likely to take the form of $f_{i} \wedge \neg f_{j} \wedge \dots \wedge \neg f_{n}$, indicating potential feature conflicts.
\end{enumerate}

When either diagonality or exclusivity exceeds a predefined hyperparameter threshold 
, the connector is designated as OR due to the heightened probability of feature conflicts. Otherwise, it is determined as AND. Both diagonality and exclusivity are computed on the client side based on their sample-based rules. During global rule aggregation, the FL server employs a majority voting mechanism from all client logical connectors to determine the optimal global connector. Note that, in most scenarios, the adopted logical connectors are the same for all FL clients. This consensus arises since the process of determining the logical connectors hinges on potential feature conflicts, which is an intrinsic quality of the features that remains consistent regardless of data distribution. Given that all clients share the same feature space, their assessments concerning potential feature conflicts are likely to align.

\subsection{Federated Rule Aggregation}
In each round of FL training, the server receives local models and local logic rules from the clients. The server first determines the appropriate logical connector following the approach described in the previous section.
It then identifies a subset of rules that optimises model performance. For a given training round, the FL server receives $K$ rules pertaining to class $c$ from $K$ FL clients. In practice, the maximum number of distinct rules the server can obtain is fewer than $K$ for several reasons. Firstly, not every client is capable of generating reliable rules. We establish an accuracy threshold to filter out rules (i.e., only models exceeding the given threshold are deemed reliable). Rules derived from models with sub-par accuracy are deemed as not credible. Secondly, even if a client model meets the accuracy threshold, it might contain biased data which does not support the predicted class $c$. Consequently, no rule is generated for class $c$. Lastly, it is possible for multiple clients to produce identical rules.

To identify the optimal combination of rules, we leverage the beam search algorithm~\cite{lowerre1976harpy}, a greedy approach commonly adopted in natural language processing and machine translation. Instead of exploring every possible sequence, beam search maintains the top $t$ sequences at each step and extends them further. This approach offers a trade-off between computational cost and solution quality. In \methodname{}, the sequences of rules are ranked based on the rule accuracy values with respect to the validation dataset on the FL server.

\subsection{FL Client Weight Assignment}
In FL, assigning appropriate weights to clients is pivotal, especially when dealing with biased or noisy datasets. Under \methodname{}, we introduce a novel method to compute client weights predicated on the logic rules from each client. The weight assigned to a client for a given training round is set to be directly proportional to the frequency with which its rules are selected by the server. Suppose in one round, client $k$ constructs $C$ rules across $C$ classes (some rules might be empty). Out of these rules, $p$ rules are aggregated into the global rule set. The weight for client $k$ in this round is:
\begin{equation}
    w_{k} = \frac{p_{k}}{\sum_{i=1}^{K}p_{i}}. \label{eq:5}
\end{equation}

If a client fails to generate any valid rules, or if its rules are not selected by the server, $w_{k}$ is set to 0. There are two reasons for a rule to be excluded from the global set: 1) the low accuracy of the rule, and 2) its integration might compromise the effectiveness of the current global rules. A high value of $w_{k}$ indicates that the client has made significant contributions in terms of rules across multiple classes in this training round, and thus shall be assigned a higher weight.

\subsection{Time Complexity Analysis}

In the rule generation process for $K$ clients, each with $N$ data points, the complexity for a single client to generate rules is $O(N)$. Considering $C$ classes, the local rule aggregation comprises rule ranking with a complexity of $O(N\log N)$ and iterative inclusion with a complexity of $O(N)$. Therefore, the overall complexity for local rule aggregation is $O(N\log N)$. During the global rule aggregation phase, beam search is employed with a beam width of $b$, leading to a worst-case complexity of $O(bK^2)$ for each class. However, since $N$ is significantly larger than $K$, the time complexity for the entire rule generation and aggregation process is $O(N\log N)$.

\section{Experimental Evaluation}
To evaluate the effectiveness of \methodname{}, we conduct experiments on four distinct datasets comparing it against three alternative approaches. Additionally, we assess \methodname{}'s performance under noisy data conditions. The outcomes are compared using three metrics: 1) model accuracy, 2) rule accuracy, and 3) rule fidelity. An ablation study is also conducted to highlight the importance of the logical connector selection method in \methodname{}.

\subsection{Experiment Settings}

Following the dataset settings~\cite{barbiero2022entropy}, we adopt four benchmark datasets: 1) MNIST(Even/Odd)~\cite{lecun1998mnist}, 2) CUB~\cite{wah2011caltech}, 3) V-Dem~\cite{coppedge2021v} and 4) MIMIC-II~\cite{saeed2011multiparameter}. MNIST and CUB are designed following the ``image $\rightarrow$ features $\rightarrow$ classes'' setting. V-Dem and MIMIC are tabular datasets that directly map input to classes. To clarify, in MNIST(Even/Odd), the dataset is augmented from the original ``image $\rightarrow$ digits'' pattern in MNIST into the ``image $\rightarrow$ digits $\rightarrow$ parity'' pattern, where the digits are extracted features and the parity is the final prediction.
For each of the datasets, we create two different data settings: 1) a centralised setting, serving as a baseline representing non-federated learning; and 2) a federated data setting~\cite{mcmahan2017communication}, which ensures a uniform distribution of data across FL clients.
Using the MNIST dataset as an illustrative example: in the federated data setting, each of the 10 clients holds a random 10\% of the entire dataset. 

In order to emulate real-world scenarios, we introduce noises into the client datasets. We choose $t\%$ of clients from the entire pool of $K$ clients, and then substitute a certain percentage of their local data with noisy data created through random label shuffling. The noise level $t\%$ is incrementally raised from 20\% to 80\% in the increment of 20\%. We then evaluate the performance of both the global model and global rules using the server's test dataset.

\begin{table*}[t]
\centering
\caption{Experiment results. We mark the best performance in bold. `-' indicates that the given evaluation metric is not applicable for an approach.}
\label{result1}
\begin{tabular}{|cl|c|ccc|ccc|}
\hline
 &     & Centralised Learning  & Distributed Decision Tree (DDT)     & FedAvg-Logic     & \methodname{} \\ \hline


\multicolumn{1}{|c|}
{\multirow{3}{*}{MNIST}} 
 & model \mbox{accuracy} & 99.84\% & 99.78\% & 99.80\% & \textbf{99.96\%} \\ 
\multicolumn{1}{|c|}{}
 & rule \mbox{accuracy} & 99.84\% & 99.71\% & 99.84\% & \textbf{99.95\%}\\ 
\multicolumn{1}{|c|}{}
 & rule \mbox{fidelity} & 99.95\% & - & 99.89\% & \textbf{99.97\%} \\ 
\hline


\multicolumn{1}{|c|}
{\multirow{3}{*}{CUB}} 
 & model \mbox{accuracy} & 82.20\% & 83.22\% & 88.64\% & \textbf{89.49\%} \\ 
\multicolumn{1}{|c|}{}
 & rule \mbox{accuracy} & 91.71\% & 87.87\% & 74.89\% &  \textbf{90.67\%}\\ 
\multicolumn{1}{|c|}{}
 & rule \mbox{fidelity} & 99.71\% & - & 98.61\% & \textbf{99.64\%} \\ 
\hline

\multicolumn{1}{|c|}
{\multirow{3}{*}{V-Dem}} 
 & model \mbox{accuracy} & 93.32\% & 92.55\% & 90.95\% & \textbf{93.08\%} \\ 
\multicolumn{1}{|c|}{}
 & rule \mbox{accuracy} & 90.84\% & 92.52\% & 86.71\% &  \textbf{93.08\%}\\ 
\multicolumn{1}{|c|}{}
 & rule \mbox{fidelity} & 94.59\% & - & 81.11\% & \textbf{100.00\%} \\ 
\hline

\multicolumn{1}{|c|}
{\multirow{3}{*}{MIMIC-II}} 
 & model \mbox{accuracy} & 76.96\% & 76.40\% & 80.89\% & \textbf{82.02\%} \\ 
\multicolumn{1}{|c|}{}
 & rule \mbox{accuracy} & 66.56\% & 67.80\% & \textbf{68.27\%} &  65.15\% \\ 
\multicolumn{1}{|c|}{}
 & rule \mbox{fidelity} & 77.24\% & - & \textbf{79.77\%} & 79.21\%\\ 
\hline

\end{tabular}%
\end{table*}

\begin{figure*}[t]
    \centering
    \begin{subfigure}{0.31\linewidth}
        \centering
        \includegraphics[width=\linewidth]{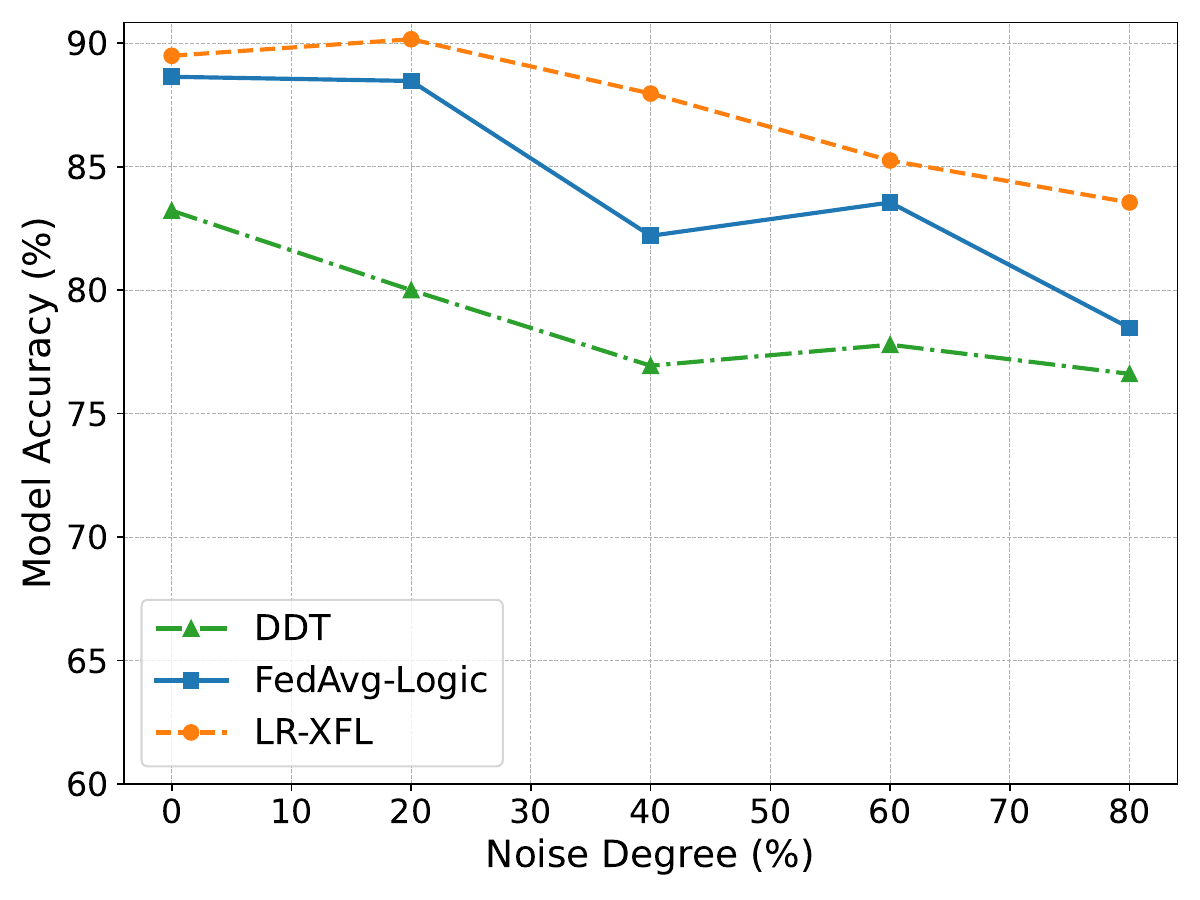}
        \caption{Model accuracy}
    \end{subfigure}%
    \hfill 
    \begin{subfigure}{0.31\linewidth}
        \centering
        \includegraphics[width=\linewidth]{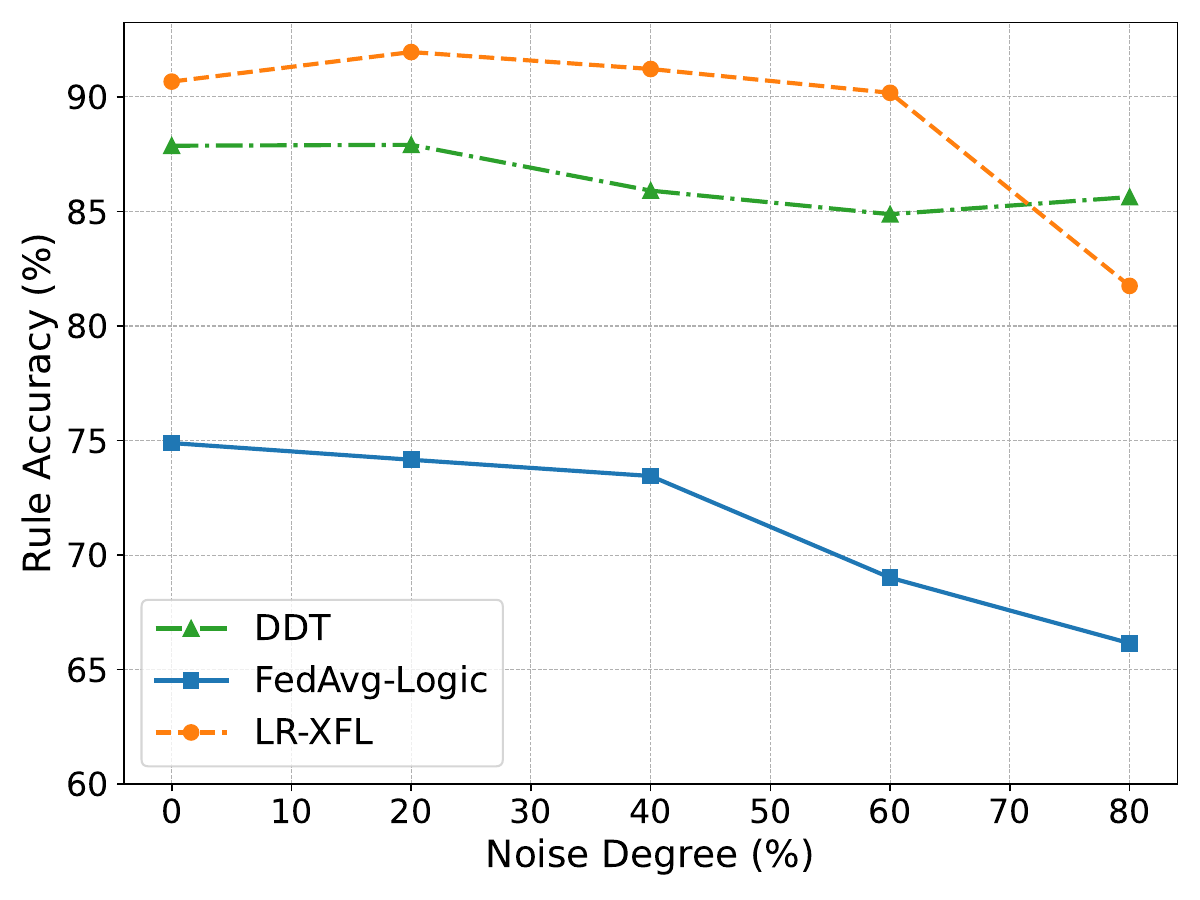}
        \caption{Rule accuracy}
    \end{subfigure}%
    \hfill
    \begin{subfigure}{0.31\linewidth}
        \centering
        \includegraphics[width=\linewidth]{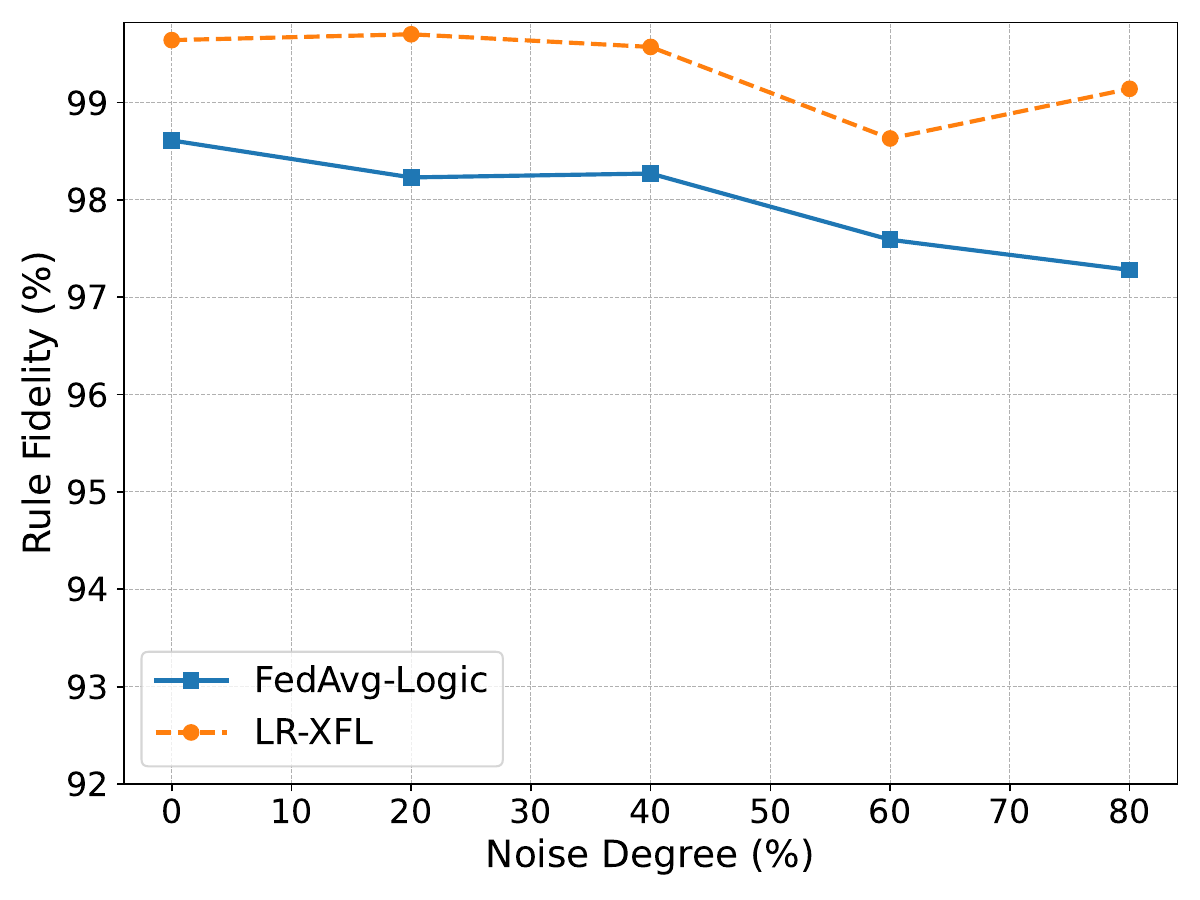}
        \caption{Rule fidelity}
    \end{subfigure}
    \caption{Experiment results under different noise level settings.}
    \label{fig:three_figs}
\end{figure*}

\subsection{Comparison Baselines}
Since there is no existing work that is specifically designed to generate and aggregate rules about FL local models and leverage such rules for FL model aggregation like \methodname{}, we compare \methodname{} against the following three baseline approaches in our experiments:
\begin{enumerate}
    \item \textbf{Centralised Learning}: This method applies the entropy-based logic explanations of neural network~\cite{barbiero2022entropy} under the centralised setting with direct access to all data. It uses the OR operator to aggregate sample-level rules into class-level rules.

    \item \textbf{Distributed Decision Tree (DDT)} \cite{quinlan1986induction}: In the DDT approach, each client maintains its own decision tree. After local training, clients share their respective decision trees with the server. The server then evaluates the received local decision trees on a validation dataset to identify the optimal tree. This top-performing local tree is subsequently used as the global decision tree for both predictions and rule generation. As decision trees are not an iterative process, clients are restricted to using the global decision tree without any further optimisation. Decision trees naturally produce rules. In our experiments, if features (considering $f_{i}$ for instance) in the rules exceed the decision tree's split value, they are represented as $f_{i}$. Conversely, features $f_{j}$ in rules that are below the split value are denoted as $\neg f_{j}$.

    \item \textbf{FedAvg-Logic}: It is an adapted version of FedAvg \cite{mcmahan2017communication} using the entropy-based network~\cite{barbiero2022entropy} as the base model. It uses the OR operator to connect for rule aggregation. Clients are assigned the same weights regardless of the performance of their rules.

\end{enumerate}

\subsection{Evaluation Metrics}
We compare the performance of \methodname{} and the baselines using the following evaluation metrics:
\begin{enumerate}
    \item \textbf{Classification Accuracy}: It is calculated as the number of correct predictions divided by the total number of predictions. This metric evaluates the extent to which a method accurately predicts the target outcomes.

    \item \textbf{Rule Accuracy}: Rule accuracy measures the consistency between the rule predictions and the ground truth labels. Let $r^c$ represent a rule of class $c$, where there are $C$ distinct classes. Suppose the ground truth marked $P$ data points to class $c$ and $Q$ data points to other classes. Among the $P$ data points, $p$ of them satisfy the propositions in the rule $r^c$, while out of the $Q$ data points, $q$ of them do not satisfy the propositions of rule $r^c$. The accuracy of this rule, denoted as $RuleAcc_c$, is defined as: 
    \begin{equation}
    RuleAcc_c = \frac{p+q}{P+Q},
    \end{equation}
    
    The overall rule accuracy $RuleAcc$ is defined as the averaged $RuleAcc_c$ across all classes.

    \item \textbf{Rule Fidelity}: Rule fidelity assesses the consistency between rule predictions and model predictions. It is computed by adapting the rule accuracy formula, where the ground truth counts $P$ and $Q$ are replaced with the model predicted counts of $P$ for class $c$ and $Q$ for other classes.

\end{enumerate}
The evaluation metrics are calculated on a test dataset stored on the server. The higher the values of these metrics, the better the performance of a given approach.

\subsection{Results and Discussion}
Table~\ref{result1} shows the comparison results between \methodname{} and the baselines. 

\subsubsection{Model Performance}
Under all experiment settings, \methodname{} achieves the highest test accuracy among the federated approaches. Compared to Centralised Learning, which serves as a reference point for non-FL settings with direct access to raw data, \methodname{} often matches or even surpasses its performance. The performance can be attributed to the effectiveness of the \methodname{} automatic logical connector selection method when applied on the client side on local sample-level rules. 
Under FL settings, \methodname{}'s advantage over FedAvg-Logic is primarily due to its client weight assignment mechanism. This mechanism effectively identifies clients whose rules make a significant contribution to the global rule set, subsequently granting them increased weights. As a result, the global server relies more heavily on these high-performing clients.
Both \methodname{} and FedAvg-Logic often outperform DDT. This can be attributed to their capability to leverage insights from multiple FL clients during the training phase. In contrast, DDT selects the local model with the best rule accuracy on the validation set, without integrating local models from other FL clients.
On average, \methodname{} achieves 1.19\% and 3.58\% higher model accuracy than FedAvg-Logic and DDT, respectively.

\subsubsection{Rule Generation Performance}
Rule accuracy and rule fidelity are critical for evaluating the interpretability and trustworthiness of a model. Under FL settings, \methodname{} consistently achieves higher or comparable performance in these respects compared to FedAvg-Logic and DDT. Large differences in rule accuracy between \methodname{} and FedAvg-Logic have been observed under the CUB and V-Dem datasets. This highlights the advantage of the automatic logical connector selection method of \methodname{}. Unlike FedAvg-Logic, which aggregates local rules with the OR logical connector, \methodname{} determines the most appropriate logical connector for any given scenario. In addition, the performance gain of \methodname{} over FedAvg-Logic can also be attributed to its rule selection and the client weight assignment mechanisms during FL model aggregation. DDT, given its inherent structure that precludes rule aggregation, occasionally outperforms FedAvg-Logic by avoiding potential issues of indiscriminate rule expansion. On average, \methodname{} achieves 5.81\% and 0.27\% higher rule accuracy than FedAvg-Logic and DDT, respectively.

The high rule fidelity achieved by \methodname{} highlights the consistency between the rules and the FL model predictions. The rule fidelity is not applicable for DDT since it is inherently rule-based. This metric, which evaluates the alignment between rule-based and model-based predictions, is always 100\% for models where rules are direct representations of the model predictions. On average, \methodname{} achieves 5.41\% higher rule fidelity than FedAvg-Logic.

\subsubsection{Resistance to Noise}
We also conduct experiments to study the robustness of DDT, FedAvg-Logic and \methodname{} against noise in FL clients' local data, taking the CUB dataset as a benchmark. Figure~\ref{fig:three_figs} illustrates the experiment results under different noise level settings. It can be observed that as the noise level increases, both model accuracy and rule accuracy decrease for all approaches. However, rule fidelity remains relatively stable, suggesting that the alignment of rules with model predictions is upheld. \methodname{} consistently achieves the highest model accuracy under various noise levels. Moreover, the degradation in its performance is moderate compared to FedAvg-Logic and DDT as the noise levels increase. Notably, the rule accuracy of \methodname{} remains relatively stable up to a noise level of 60\%. This can be attributed to its rule selection and client weight assignment mechanism, which guards the global FL model against integrating information from noisy clients. Interestingly, the DDT model achieves the best performance when the noise level reaches 80\%. This can be attributed to its strategy of only adopting the best single client model as the global decision tree, maximally mitigating the exposure to noise at the expense of limiting knowledge sharing among FL clients.

\subsection{Ablation Study}


\begin{table}[t]
    \centering
    \caption{Ablation study results.} \label{tab:ablation}
    \begin{tabular}{|r|c|c|}
        \hline
         & \methodname{} (ablated) & \methodname{} \\\hline
          Model Accuracy & 87.96\% & \textbf{89.49\%} \\\hline
          Rule Accuracy & 76.29\% & \textbf{90.67\%}  \\\hline
          Rule Fidelity & 98.83\% & \textbf{99.64\%} \\\hline
    \end{tabular}
\end{table}

We conduct ablation studies to evaluate the impact of the automatic logical connector selection method on the performance of \methodname{}. In the experiments, threshold values for diagonality and exclusivity were set to 0.9 and 0.8, respectively, based on hyperparameter tuning. Since the logical connector chosen by \methodname{} for the CUB dataset is AND, we created an ablated version of \methodname{} using OR as the logical connector for rules under the CUB dataset. The results are shown in Table~\ref{tab:ablation}. It can be observed that the ablated version of \methodname{} achieves a lower model accuracy and rule fidelity, and a pronounced decline in rule accuracy compared to the full version of \methodname{}. The results demonstrate that using OR to connect rules can erroneously reduce global rule accuracy when rules may contain complementary information. This underscores the importance of the proposed automatic logical connector selection method.

\section{Conclusions}
In this paper, we proposed \methodname{}, a first-of-its-kind logic-based explainable federated learning framework. It is capable of deriving accurate global rules from local rules without requiring access to clients' local data. The most appropriate logical connectors for aggregating client rules are automatically determined by \methodname{} based on the characteristics of clients' local data. This novel design significantly enhances the trustworthiness of the resulting model. Moreover, the aggregated rules play a pivotal role in determining client weights during FL model aggregation. This transparent design enables domain experts to engage actively in the validation, refinement and adjustment of the rules, thereby helping improve the result FL model as well.


\section{Acknowledgements}
This research/project is supported, in part, by the National Research Foundation Singapore and DSO National Laboratories under the AI Singapore Programme (AISG Award No: AISG2-RP-2020-019); the RIE 2020 Advanced Manufacturing and Engineering (AME) Programmatic Fund (No. A20G8b0102), Singapore; and the Joint NTU-WeBank Research Centre on FinTech, Nanyang Technological University, Singapore.

\bibliography{aaai24}

\clearpage
\appendix

\section{Appendix}

\subsection{Dataset Description}
In federated learning (FL) settings, the dataset is partitioned to facilitate both server-based and client-based operations. Specifically, 20\% of the data is allocated to the server: 10\% for the validation of user selection based on their generated rules and the remaining 10\% for testing the aggregated global rules and assessing global rule accuracy. For our experiments, we set up 10 FL clients, with each client holding 10\% of total client data. Within each client's data portion, a split of 90\% for training, 5\% for validation, and 5\% for testing is adopted.

Building on the dataset configurations presented in~\cite{barbiero2022entropy}, we utilised four benchmark datasets: 
\begin{enumerate}
    \item MNIST(Even/Odd)~\cite{lecun1998mnist},
    \item CUB~\cite{wah2011caltech},
    \item V-Dem~\cite{coppedge2021v}, and 
    \item MIMIC-II~\cite{saeed2011multiparameter}.
\end{enumerate}

\textbf{MNIST(Even/Odd)} is a derivative of the MNIST dataset~\cite{lecun1998mnist}, following the ``image $\rightarrow$ features $\rightarrow$ classes'' pattern. Comprising 60,000 instances of 28x28 pixel handwritten digits, its features are the digits 0-9 predicted by a ResNet10 model~\cite{he2016deep}. The ultimate classification task focuses on determining the parity of the digit, i.e., even or odd.

\textbf{CUB} (Caltech-UCSD Birds-200-2011) adheres to the ``image $\rightarrow$ features $\rightarrow$ classes" pattern. It features 11,788 images spanning 200 bird subcategories. Each image is annotated with 312 binary attributes capturing characteristics like colour, size, shape, and pattern. Based on the preprocessing from~\cite{koh2020concept}, we include 108 of these features in our study. The features for CUB are derived using the ResNet10~\cite{he2016deep}, and the ultimate classification pertains to the 200 bird subcategories.

\textbf{V-Dem} (Varieties of Democracy) is based on the ``input $\rightarrow$ classes" pattern. This tabular dataset encompasses latent regime characteristics at various granularity levels. Our study, aligned with the settings in~\cite{barbiero2022entropy}, works with 3,743 data points post-processing. The dataset retains 14 human-understandable features guiding the predictions for two democracy classes: electoral and non-electoral.

\textbf{MIMIC-II} (Multiparameter Intelligent Monitoring in Intensive Care II) also employs the ``input $\rightarrow$ classes" pattern, offering a comprehensive view into intensive care practices. This dataset captures physiological signals, vital stats, and extensive clinical data from ICU patients. With preprocessing aligned to~\cite{barbiero2022entropy}, our analysis incorporates 1,776 data points, each embedded with 90 features. The classification task is to predict patient recovery outcomes post-ICU admission.

\subsection{Rules Generated by \methodname{}}

Table~\ref{rule_FL_cub} displays the global rule and its accuracy on the server test dataset for the Swainson Warbler class, as determined during the final federated training round. It also presents each client's rule and their respective accuracies on this test dataset. Notably, the rules from clients 2, 4, 8, and 9 have been integrated to construct the global rule. By aggregating client local rules using AND logic operator, the global rule achieves an accuracy of 100.00\%, surpassing that of any individual local rule.

Table~\ref{rule_mnist} presents the global rule, rule accuracy, and rule fidelity for distinguishing between even and odd numbers in the MNIST(Even/Odd) dataset. Though the logical connector for this dataset is decided as OR by~\methodname{},  it is not evident in the rules. This omission arises because a particular sample-level rule attained optimal accuracy. Thus, during class-level aggregation within clients and global-level aggregation across clients, the inclusion of additional rules with this optimal sample-level rule does not enhance accuracy further. This sample-level rule ultimately becomes the final global rule.

Table~\ref{rule_cub} displays the global rule, rule accuracy, and rule fidelity for 200 bird subcategories in CUB dataset. The global logical connector in the CUB dataset is AND.

Table~\ref{rule_vdem} showcases the global rules, their accuracy, and fidelity concerning the presence or absence of electoral democracy in the V-Dem dataset. The logical connector employed in the V-Dem dataset is AND.

Table~\ref{rule_mimic} outlines the global rules, rule accuracy, and rule fidelity for the recovery status of patients in the MIMIC-II dataset. The MIMIC-II dataset adopts AND as its primary logical connector.

\onecolumn
\centering
\begin{longtable}{l|p{.75\linewidth}|l}
\caption{Global and local rules for the Swainson Warbler in CUB. Clients contributing to the global rule are highlighted in bold. A '-' in the rule column signifies that the respective client did not produce a rule for this class.}
\label{rule_FL_cub}\\

\hline
& Rule & Accuracy \\
\hline
Global & $\text{has\_bill\_shape-all-purpose} \wedge \text{has\_tail\_shape-notched\_tail} \wedge \text{has\_crown\_color-brown} \wedge \neg \text{has\_throat\_color-white} \wedge \neg \text{has\_bill\_length-shorter\_than\_head}$ & 100.00\% \\
\hline
Client 1 & $\text{has\_under\_tail\_color-brown} \wedge \neg \text{has\_bill\_length-shorter\_than\_head}$ & 55.24\% \\
\hline
\textbf{Client 2} & $\text{has\_tail\_shape-notched\_tail} \wedge \neg \text{has\_bill\_length-shorter\_than\_head}$ & 66.50\% \\
\hline
Client 3 & $\text{has\_bill\_shape-all-purpose} \wedge \text{has\_back\_color-brown} \wedge \neg \text{has\_underparts\_color-brown}$ & 52.81\% \\
\hline
\textbf{Client 4} & $\text{has\_tail\_shape-notched\_tail} \wedge \neg \text{has\_bill\_length-shorter\_than\_head}$ & 66.50\% \\
\hline
Client 5 & $\text{has\_tail\_shape-notched\_tail} \wedge \neg \text{has\_bill\_shape-cone} \wedge \neg \text{has\_upperparts\_color-grey} \wedge \neg \text{has\_under\_tail\_color-black} \wedge \neg \text{has\_primary\_color-yellow}$ & 54.53\% \\
\hline

\end{longtable}

\centering
\begin{longtable}{p{.78\linewidth}|l|l}
\caption{Global rules for the 2 classes in the MNIST(Even/Odd) dataset, along with their respective accuracy and fidelity on the global test set.}
\label{rule_mnist}\\

\hline
Rule & Accuracy & Fidelity \\\endhead
\hline
$ \text{Odd} \leftrightarrow \neg0 \wedge \neg2 \wedge \neg4 \wedge \neg6 \wedge \neg8 $ & 99.95\% & 99.97\%\\
\hline
$ \text{Even} \leftrightarrow \neg1 \wedge \neg3 \wedge \neg5 \wedge \neg7 \wedge \neg9 $ & 99.97\% & 99.98\%\\
\hline
\end{longtable}

\begin{longtable}{p{.78\linewidth}|l|l}
\caption{Global rules for the 2 classes in the V-Dem dataset, along with their respective accuracy and fidelity on the global test set.}
\label{rule_vdem}\\

\hline
Rule & Accuracy & Fidelity \\\endhead
\hline
$ \text{Electoral\_democracy} \leftrightarrow \text{v2x\_freexp\_altinf} \wedge \text{v2xel\_frefair} \wedge \text{v2x\_elecoff} $ & 93.08\% & 100.00\%\\
\hline
$ \text{Non\_electoral\_democracy} \leftrightarrow \neg \text{v2xel\_frefair} $ & 93.08\% & 100.00\%\\
\hline
\end{longtable}

\begin{longtable}{p{.78\linewidth}|l|l}
\caption{Global rules for the 2 classes in the MIMIC-II dataset, along with their respective accuracy and fidelity on the global test set.}
\label{rule_mimic}\\

\hline
Rule & Accuracy & Fidelity \\\endhead
\hline
$ \text{Recover} \leftrightarrow \neg \text{stroke\_flg} \wedge \neg \text{mal\_flg} \wedge \neg \text{age\_HIGH} $ & 78.33\% & 76.40\%\\
\hline
$ \text{Non\_recover} \leftrightarrow \text{resp\_flg} \wedge \text{age\_HIGH} \wedge \text{hour\_icu\_intime\_LOW} \wedge \neg \text{age\_LOW} \wedge \neg \text{chloride\_first\_LOW} \wedge \neg \text{map\_1st\_HIGH} \wedge \neg \text{po2\_first\_HIGH} \wedge \neg \text{sapsi\_first\_LOW} $ & 51.98\% & 82.02\%\\

\hline
\end{longtable}

\centering
\begin{longtable}{p{.78\linewidth}|l|l}
\caption{Global rules for the 200 classes in the CUB dataset, along with their respective accuracy and fidelity on the global test set.}
\label{rule_cub} \\

\hline
Rule & Accuracy & Fidelity \\\endhead

\hline
$ \text{Black\_footed\_Albatross} \leftrightarrow \text{has\_bill\_shape-hooked\_seabird} \wedge \neg \text{has\_underparts\_color-white }$ & 73.33\% & 97.97\% \\
\hline
$ \text{Laysan\_Albatross} \leftrightarrow \text{has\_bill\_shape-hooked\_seabird} \wedge \neg \text{has\_size-medium\_9\_-\_16\_in} \wedge \neg \text{has\_crown\_color-black }$ & 94.40\% & 99.83\% \\
\hline
$ \text{Sooty\_Albatross} \leftrightarrow \text{has\_upper\_tail\_color-grey} \wedge \text{has\_back\_pattern-solid} \wedge \text{has\_bill\_color-black} \wedge \neg \text{has\_size-small\_5\_-\_9\_in} \wedge \neg \text{has\_primary\_color-grey }$ & 100.00\% & 100.00\% \\
\hline
$ \text{Groove\_billed\_Ani} \leftrightarrow \text{has\_head\_pattern-plain} \wedge \text{has\_bill\_length-shorter\_than\_head} \wedge \neg \text{has\_bill\_shape-all-purpose} \wedge \neg \text{has\_bill\_shape-cone} \wedge \neg \text{has\_underparts\_color-white} \wedge \neg \text{has\_bill\_length-about\_the\_same\_as\_head} \wedge \neg \text{has\_nape\_color-white} \wedge \neg \text{has\_wing\_shape-rounded-wings} \wedge \neg \text{has\_shape-perching-like }$ & 95.41\% & 98.98\% \\
\hline
$ \text{Crested\_Auklet} \leftrightarrow \text{has\_underparts\_color-grey} \wedge \text{has\_breast\_color-grey} \wedge \neg \text{has\_eye\_color-black} \wedge \neg \text{has\_shape-perching-like }$ & 100.00\% & 99.83\% \\
\hline
$ \text{Least\_Auklet} \leftrightarrow \text{has\_breast\_pattern-striped} \wedge \neg \text{has\_shape-perching-like }$ & 100.00\% & 100.00\% \\
\hline
$ \text{Parakeet\_Auklet} \leftrightarrow \text{has\_primary\_color-white} \wedge \text{has\_leg\_color-grey} \wedge \neg \text{has\_eye\_color-black} \wedge \neg \text{has\_shape-perching-like} \wedge \neg \text{has\_bill\_color-black }$ & 100.00\% & 99.83\% \\
\hline
$ \text{Rhinoceros\_Auklet} \leftrightarrow \text{has\_leg\_color-buff} \wedge \neg \text{has\_shape-perching-like }$ & 100.00\% & 100.00\% \\
\hline
\end{longtable}

\end{document}